\documentclass[11pt,a4paper]{article}
\usepackage[hyperref]{emnlp2020}
\usepackage{times}
\usepackage{latexsym}
\usepackage{arydshln}

\usepackage{url}
\usepackage{color,soul}
\usepackage{graphicx}
\usepackage{multirow}
\usepackage{multicol}
\usepackage{amsmath,amssymb,amsfonts}
\usepackage{pifont}
\usepackage[capitalize]{cleveref}
\usepackage{subcaption}

\usepackage{microtype}

\aclfinalcopy 
\def\aclpaperid{476} 

\DeclareMathOperator{\softmax}{softmax}
\DeclareMathOperator{\FC}{FC}
\DeclareMathOperator{\ReLU}{ReLU}
\DeclareMathOperator{\KL}{KL}

\newcommand{\cmark}{\ding{51}}
\newcommand{\xmark}{\ding{55}}
\newcommand{\stack}[2]{$\substack{\text{#1}\\\text{#2}}$}

\newcommand\blfootnote[1]{%
  \begingroup
  \renewcommand\thefootnote{}\footnote{#1}%
  \addtocounter{footnote}{-1}%
  \endgroup
}

\title{MIME: MIMicking Emotions for Empathetic Response Generation}

\author{Navonil Majumder$^\dagger$,
  Pengfei Hong$^\dagger$,
  Shanshan Peng$^{\dagger\ast}$,
  Jiankun Lu$^{\dagger\ast}$,\\
  \textbf{Deepanway Ghosal$^\dagger$,
  Alexander Gelbukh$^\diamond$,
  Rada Mihalcea$^\triangle$,
  Soujanya Poria$^\dagger$}\\\\
  
  $^\dagger$ Singapore University of Technology and Design, Singapore\\
  $^\diamond$ CIC, Instituto Polit\'ecnico Nacional, Mexico\\
  $^\triangle$ University of Michigan, USA\\
  \texttt{\{navonil\_majumder, sporia\}@sutd.edu.sg},\\ \texttt{\{shanshan\_peng, jiankun\_liu\}@mymail.sutd.edu.sg},\\ \texttt{\{pengfei\_hong, deepanway\_ghosal\}@mymail.sutd.edu.sg},\\ \texttt{gelbukh@cic.ipn.mx, mihalcea@umich.edu} 
  }

\date{}

\begin{document}
\maketitle
\begin{abstract}
Current approaches to empathetic response generation view the set of emotions expressed in the input text as a flat structure, where all the emotions are treated uniformly. We argue that empathetic responses often mimic the emotion of the user to a varying degree, depending on its positivity or negativity and content. We show that the consideration of this polarity-based emotion clusters and emotional mimicry results in improved empathy and contextual relevance of the response as compared to the state-of-the-art. Also, we introduce stochasticity into the emotion mixture that yields emotionally more varied empathetic responses than the previous work. We demonstrate the importance of these factors to empathetic response generation using both automatic- and human-based evaluations. The implementation of MIME is publicly available at \url{https://github.com/declare-lab/MIME}.
\blfootnote{$^\ast$ signifies equal contribution}
\end{abstract}

\section{Introduction}
\label{intro}

Empathy is a fundamental human trait that reflects our ability to understand and reflect the thoughts and feelings of the people we interact with. In the social sciences, research on empathy has evolved into an entire field of study, addressing the social underpinning of empathy \cite{Singer09The}, the cognitive and emotion aspects of empathy \cite{Smith06Cognitive}, and its connection to personal and demographic traits \cite{Dymond50Personality,Eisenberg14,Krebs75Empathy}. The study of empathy has found a wide range of applications in healthcare, including psychotherapy \cite{Bohart97Empathy} or more broadly as a mechanism to improve the quality of care \cite{Mercer02Empathy}.

Computational models of empathy have been proposed only in recent years, partly because of the complexity of this behavior which makes it difficult to emulate with computational approaches. In natural language processing, the methods proposed to date address the tasks of understanding expressions of empathy in newswire \cite{Buechel18Modeling}, counseling conversations \cite{Perez-Rosas17Understanding}, or generating empathy in dialogue  \cite{Shen20Counseling,lin-etal-2019-moel}. Work has also been done on the construction of empathy lexicons \cite{Sedoc20Learning} or large empathy dialogue datasets \cite{rashkin2018empathetic}.

\begin{figure}
    \centering
    \includegraphics[width=\linewidth]{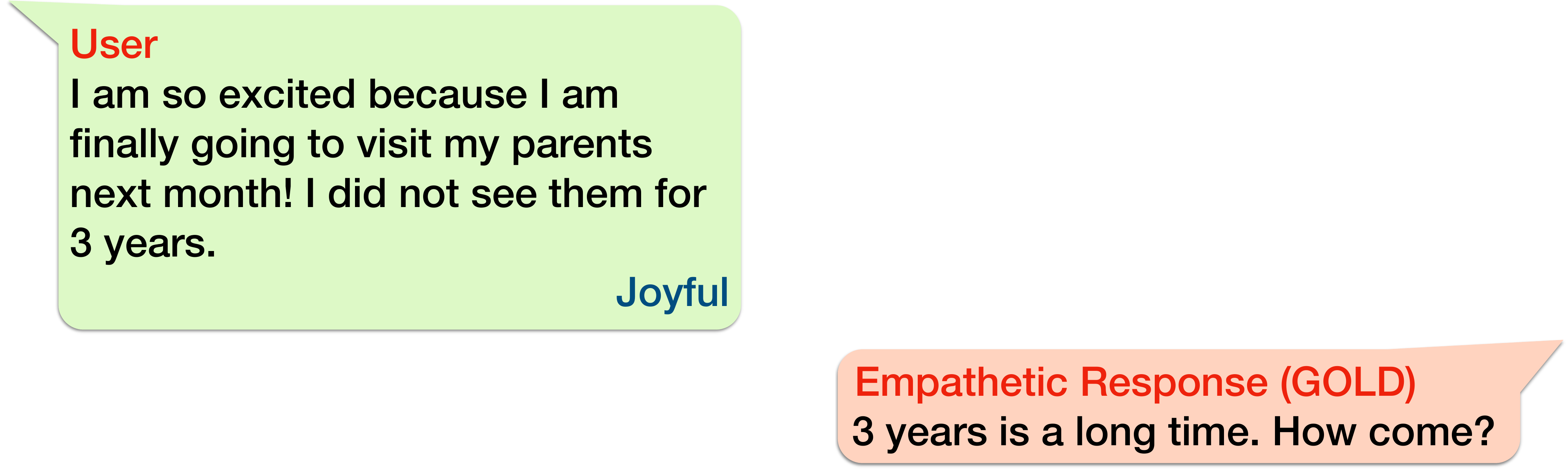}
    \caption{An instance where a positive context is responded with ambivalence.}
    \label{fig:ambivalence}
\end{figure}

In this paper, we address the task of generating empathetic responses that mimic the emotion of the speaker while accounting for their affective charge (positive or negative). We adopt the idea of emotion mixture, as the state-of-the-art MoEL~\cite{lin-etal-2019-moel}, to achieve the appropriate balance of emotions in positive and negative emotion groups. However, inspired by \citet{vhred}, we introduce stochasticity into the mixture at emotion-group level for varied responses. This becomes particularly important in cases where the input utterance can be responded with ambivalent, yet befitting utterances. \cref{fig:ambivalence} shows one such example where the response to a positive utterance is ambivalent. 

The paper makes two important contributions. First, it introduces a new approach for empathetic generation that encodes context and emotions, and uses emotion stochastic sampling and emotion mimicry to generate responses that are appropriate and empathetic for positive or negative statements. We show that this approach leads to performance exceeding the state-of-the-art when trained and evaluated on a large empathy dialogue dataset. Second, through extensive feature ablation experiments, we shed light on the role played by emotion mimicry and emotion grouping for the task of empathetic response generation.
\begin{figure*}[ht!]
    \centering
    \includegraphics[width=0.95\linewidth]{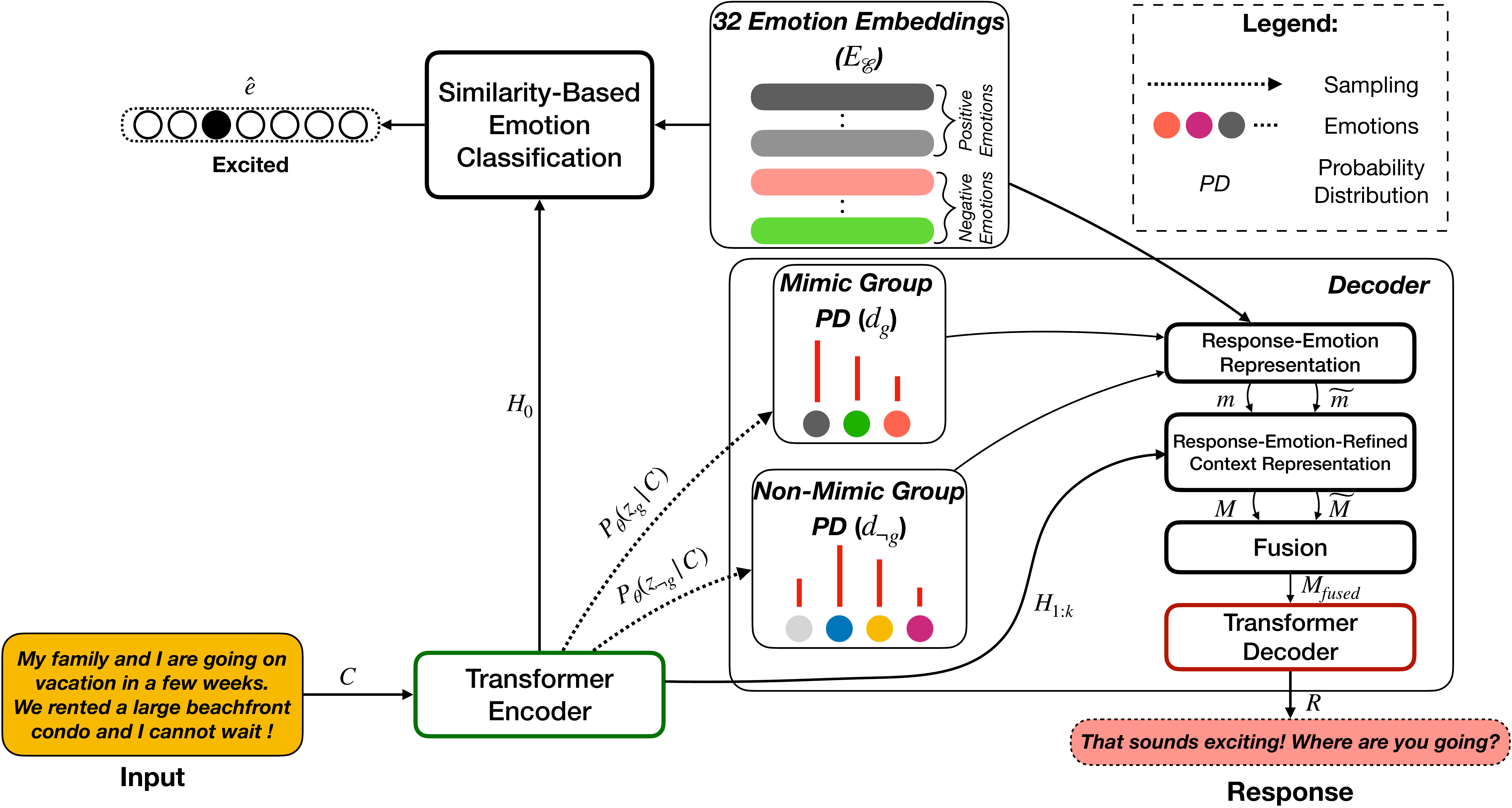}
    \caption{Architecture of our model (MIME).}
    \label{fig:model}
\end{figure*}

\section{Related Work}
Open domain conversational models have made good progress in recent years \cite{serban2016generative, vinyals2015neural, wolf2019transfertransfo}. 
Many of them can generate persona-consistent \cite{zhang-etal-2018-personalizing}
and diverse \cite{cai2018skeletontoresponse}
responses, but those are not necessarily empathetic. 

Producing empathetic responses requires apt handling of emotions and sentiments \cite{fung-etal-2016-zara-supergirl,Winata2017NoraTE,bertero-etal-2016-real}. \citet{zhou2018emotional} model psychological concepts as memory states in LSTM \cite{HochSchm97} and employ emotion-category embeddings in the decoding process. \citet{ijcai2018-618} presents a GAN \cite{goodfellow2014generative} based framework with emotion-specific generators. On a larger scale, \cite{zhou-wang-2018-mojitalk} use the emojis in Twitter posts as emotion labels and introduce an attention-based \cite{luong2015effective} Seq-to-Seq \cite{sutskever2014sequence} model with Conditional Variational Autoencoder \cite{sohn2015learning} for emotional response generation. 
However, they only produce affective responses with user-provided emotion, which may not necessarily be empathetic to the speakers.
\citet{wu2019simple} introduce a dual-decoder network to generate responses with given sentiment (positive or negative). 
\citet{inproceedings} formulate a reinforcement learning problem to maximize user's sentimental feeling towards the generated response. \citet{lin-etal-2019-moel} present an encoder-decoder model with each emotion having a dedicated decoder. 


Variational Bayes \cite{kingma2013auto,rezende2014stochastic}
has been widely adopted into natural language generation tasks \cite{bowman2015generating} and successfully extended to dialog generation tasks 
\cite{vhred}.
The prominent approach by Hierarchical Encoder-Decoder (VHRED) \cite{vhred} integrates VAE with the sequence-to-sequence decoder based on Markov assumptions.

\section{Methodology}

Our model MIME is based on the assumption that empathetic responses often mimic the emotion of the speaker~\cite{Carr5497} --- in our case, the human subject or user. For example, positively-charged utterances are usually responded with positive emotions, although they can also be ambivalent as illustrated in \cref{fig:ambivalence}. On the other hand, responding to negatively-charged utterances often requires composite emotions that agree with the user's emotion, but also tries to comfort them with some positivity, such as hopefulness or silver lining. As such, we strive to balance the mimicry of context/user emotion during empathetic response generation.

To this end, we first obtain context representation using a transformer encoder architecture~\cite{NIPS2017_7181}. Similar to the state-of-the-art (SOTA) model MoEL~\cite{lin-etal-2019-moel}, we enforce emotion understanding in the context representation by classifying user emotion  during training. 
For the response emotion, we first group the 32 emotions into two groups containing \emph{positive} and \emph{negative} emotions (\cref{sec:response}). Next, a probability distribution of emotions is sampled for each of these groups that corresponds to the emotion of the response. Positive and negative response emotion representations are formed from these distributions and emotion embeddings. These two representations are appropriately combined to balance the two kinds of emotions to form the emotional representation that drives the emotional state during response generation using transformer decoder~\cite{NIPS2017_7181}. \cref{fig:model} shows the  architecture of our model.

\subsection{Task Definition}

Given the context utterances $[u_0, u_1,\dots, u_{n-1}]$, where utterance $u_i=[w_0^i, w_1^i,\dots,w_{m-1}^i]$ consists of maximum $m$ words, the task is to generate an empathetic response to the last utterance $u_{n-1}$, which is always from the target speaker or user. All the even-numbered ($u_0, u_2,\dots$) and odd-numbered ($u_1, u_3,\dots$) utterances belong to the user and the empathetic agent, respectively. Optionally, the context/user emotion $e$ can be predicted for emotion understanding. The emotions are listed in \cref{tab:emo_group}.

\subsection{Context Encoding}

Following the MoEL system \cite{lin-etal-2019-moel}, firstly, the contextual utterances are sequentially concatenated into a string of $k$ ($\leq mn$)  words $C=[u_0\oplus u_1\oplus\dots\oplus u_{n-1}]=[w_0^0, w_1^0,\dots,w_0^1,w_1^1,\dots,w^{n-1}_{m-1}]$, where $\oplus$ is the concatenation operator.

As in MoEL, each word $w^i_j$ is represented as a sum of three embeddings ($E_C$): semantic word embedding ($E_W$), positional embedding ($E_P$), and speaker embedding ($E_S$), where $E_W(w^i_j), E_P(w^i_j), E_S(w^i_j)\in \mathbb{R}^{D_{emb}}$. Therefore, the context $C$ is represented as
\begin{flalign}
E_C(C)=E_W(C)+E_P(C)+E_S(C),
\end{flalign}
where $E_C(C)\in \mathbb{R}^{k\times D_{emb}}$.

Also as in MoEL, a transformer encoder~\cite{NIPS2017_7181} is used for context propagation within the utterances and words in $C$. Moreover, inspired by BERT~\cite{devlin-etal-2019-bert}, one additional token $CTX$ is prepended to the context sequence $C$ to encode the entirety of the context:
\begin{flalign}
H&=\text{TR}_\text{Enc}^{ctx}(E_C([CTX] \oplus C])), \label{eqn:H}
\end{flalign}
where $\text{TR}_\text{Enc}^{ctx}$ is the transformer encoder of output size $D_h$ and $H\in \mathbb{R}^{(k+1)\times D_h}$ contains the context-enriched representations of the contextual words. A context-enriched representation of the $CTX$ token, $c$, is taken as the overall context representation:
\begin{flalign}
c=H_0. \label{eqn:c}
\end{flalign}

\paragraph{Emotion Embedding and Classification.} As in MoEL and also as in \citet{Rashkin2018IKT}, to explicitly infuse emotion into the context representation $c$, we train a emotion classifier on $c$. We train emotion embeddings $E_\mathcal{E}\in \mathbb{R}^{n_{emo}\times D_h}$ ($n_{emo}=32$ is the number of emotion classes) to represent each emotion. We maximize the similarity between $c$ and the user-emotion representation $E_\mathcal{E}(e)$, $e$ being the user-emotion label, using cross-entropy loss $L_{cls}$:
\begin{flalign}
s&=E_\mathcal{E}W_\mathcal{E} c^T ,\\
\mathcal{P}&=\softmax(s), \label{eqn:emo-classifier}\\
L_{cls} &= -\log \mathcal{P}[e],
\end{flalign}
where $W_\mathcal{E}\in \mathbb{R}^{D_h\times D_h}$ and $s,\mathcal{P}\in \mathbb{R}^{n_{emo}}$.

\subsection{Response Generation (Decoder)}
\label{sec:response}

Our primary assumption behind this model is that the empathetic agent mimics the user's emotion to some degree during response. Specifically, positive emotion is often responded with closely positive response. Negative emotion, however, is likely responded with negativity mixed with some positivity to uphold the moral.

\paragraph{Emotion Grouping.} We split the 32 emotion types into two groups containing 13 \emph{positive} and 19 \emph{negative} emotions, as listed in \cref{tab:emo_group}. This split is guided by our intuition.

\begin{table}[ht!]
    \centering
    \small
    \resizebox{\linewidth}{!}{
    \begin{tabular}{p{0.50\linewidth}|p{0.50\linewidth}}
        \hline
         \multicolumn{1}{c|}{Positive} & \multicolumn{1}{c}{Negative} \\
         \hline
         confident, joyful, grateful, impressed, proud, excited, trusting, hopeful, faithful, prepared, content, surprised, caring 
         &
         afraid, angry, annoyed, anticipating, anxious, apprehensive, ashamed, devastated, disappointed, disgusted, embarrassed, furious, guilty, jealous, lonely, nostalgic, sad, sentimental, terrified \\
         \hline
    \end{tabular}
    }
    \caption{32 emotions are split into two groups by emotional positivity and negativity.}
    \label{tab:emo_group}
\end{table}

\paragraph{Response-Emotion Sampling.} There is more than one correct way to respond empathetically. However, we observed that the SOTA model, MoEL, often resorts to generic and repetitive, although empathetic, responses. Therefore, inspired by \citet{vhred}, we introduce stochasticity in the response-emotion determination that results in emotionally more varied responses. In \cref{tab:with-STC vs without-STC}, we present responses generated by MIME with and without stochasticity. To this end, we sample response-emotion distributions $d_{pos}$ and $d_{neg}$, from the context $C$ --- specifically, $c$ in $\cref{eqn:c}$ ---, for both positive and negative emotion groups, respectively. Hence, we sample an unnormalized distribution $z_g$ ($g\in \{pos, neg\}$) from distribution $P_\theta(z_g|C)$. This $z_g$ is passed to a fully-connected layer ($\FC_{d_g}$) with softmax activation to obtain the normalized distribution $d_g\in \mathbb{R}^{n_g}$ ($n_{pos}=13$ and $n_{neg}=19$):
\begin{flalign}
P_\theta(z_g|C) &= \mathcal{N}(\mu_{g}^\text{prior}(C), \sigma_{g}^\text{prior}(C)),\\
z_{g} &\sim P_\theta(z_g|C),\\
d_{g} &= \softmax(\FC_{d_{g}}(z_{g})).
\end{flalign}
The emotion representation for each emotion group, $e_g\in \mathbb{R}^{D_h}$, is obtained by pooling the corresponding emotion embeddings using the respective distribution $d_g$:
\begin{flalign}
e_{g} &= d_{g} E_{\mathcal{E}_{g}}, \label{eqn:e_g}
\end{flalign}
where $E_{\mathcal{E}_g}\in \mathbb{R}^{n_g\times D_h}$ are emotion embeddings in the emotion group $g$ --- as defined in \cref{tab:emo_group}.

Sampling from distribution $P_\theta(z_g|C)$ is reparameterized as follows:
\begin{flalign}
c'&=\ReLU(\FC_{sample}(c)),\\
\mu_{g}^\text{prior}(C) &= \FC_{\mu_{g}}(c'),\\
\sigma_{g}^\text{prior}(C) &= \exp(0.5 \FC_{\sigma_{g}}(c')),\\
r &\sim \mathcal{N}(0, I),\\
z_{g} &= \mu_{g}^\text{prior}(C) + r \odot \sigma_{g}^\text{prior}(C),
\end{flalign}
where $g\in \{pos, neg\}$, $\FC_*$ are fully-connected layers with output sizes $D_h$. Following \citet{vhred}, $P_\theta(z_g|C)$ is obtained by maximizing the evidence lower-bound ($-L_g^{ELBO}$):
\begin{multline}
L_{g}^{ELBO} = \KL[Q_\psi(z_g|e_g, C) || P_\theta(z_g|C)] \\- \mathbb{E}_{Q_\psi(z_g|e_g, C)}[\log P_\theta(e_g|z_g, C)],
\end{multline}
where $Q_\psi(z_g|e_g, C)$ is the approximate posterior distribution, defined as:
\begin{multline}
Q_\psi(z_g|e_g, C) \\ 
=\mathcal{N}(\mu_{g}^\text{posterior}(e_g, C), \sigma_{g}^\text{posterior}(e_g, C)),
\end{multline}
which is similarly reparameterized, for sampling during the training only, as $P_\theta(z_g|C)$, except $e_g$ is concatenated to $c$. 

\paragraph{Emotion Mimicry.} Following \citet{Carr5497}, it is reasonable to assume that the empathetic response to an emotional utterance would likely mimic the emotion of the user to some degree. Responding empathetically to positive utterances usually requires positivity, occasionally including  ambivalence~(\cref{fig:ambivalence}). On the other hand, the responses to negative utterances should contain some empathetic negativity, but mixed with some positivity to soothe the user's negativity. Thus, we generate two distinct response-emotion-refined context representations --- mimicking and non-mimicking --- that are appropriately merged to obtain response-decoder input.

Naturally, mimicking and non-mimicking emotion representations --- $m$ and $\widetilde{m}$ ---  are defined as follows:
\begin{flalign}
m &= e_{pos}\text{ if }e\text{ is positive, otherwise }e_{neg}, \label{eqn:m} \\
\widetilde{m} &= e_{neg}\text{ if }e\text{ is positive, otherwise }e_{pos}. \label{eqn:nm}
\end{flalign}

 Firstly, response-emotion representations --- $m$ and $\widetilde{m}$ --- are concatenated to the context-enriched word representations in $H_{1:k}$ (\cref{eqn:H}) to provide the context ($C$) the cues on the response emotion:
\begin{flalign}
H_{resp} &= [H_i\oplus m]^{k}_{i=1}, \label{eqn:Hm}\\ 
\widetilde{H}_{resp} &= [H_i\oplus \widetilde{m}]^{k}_{i=1}, \label{eqn:Hnm}
\end{flalign}
where $H_{resp}, \widetilde{H}_{resp}\in \mathbb{R}^{k\times 2D_h}$ are fed to a transformer encoder ($\text{TR}_\text{Enc}^{resp}$) to obtain mimicking and non-mimicking response-emotion-refined context representations $M$ and $\widetilde{M}$, respectively:
\begin{flalign}
M &= \text{TR}_\text{Enc}^{resp}(H_{resp}),\\
\widetilde{M} &= \text{TR}_\text{Enc}^{resp}(\widetilde{H}_{resp}),
\end{flalign}
where $M, \widetilde{M}\in \mathbb{R}^{k\times D_h}$.

\paragraph{Response-Emotion-Refined Context Fusion.}

Enabling a mixture of positive and negative emotions could lead to diverse response generation as compared to considering exclusively positive or negative emotions. To achieve this mixture, we concatenate $M$ and $\widetilde{M}$ at word level, as opposed to sequence level, to obtain $M'\in \mathbb{R}^{k\times 2D_h}$. Then, $M'$ is fed to a gate consisting of a fully-connected layer ($\FC_{contrib}$) with sigmoid activation, resulting $M_{contrib}$ that determines the contribution of postive and negative response-emotion-refined contexts to the response to be generated. Subsequently, $M'$ is multiplied with the gate output, yielding the refined context $M_{adjust}$ that is fed to another fully-connected layer $\FC_{fused}$ to obtain the fused response-emotion-refined context $M_{fused}\in \mathbb{R}^{k\times D_h}$:
\begin{flalign}
M' &= [M_i\oplus \widetilde{M}_i]_{i=0}^{k-1},\\ 
M_{contrib} &= \sigma(\FC_{contrib}(M')),\\
M_{adjust} &= M_{contrib} \odot M',\\
M_{fused} &= \FC_{fused}(M_{adjust}).
\end{flalign}

\paragraph{Response Decoding.}

For the final response generation from the response-emotion-refined context $M_{fused}$, following  MoEL, a transformer decoder ($\text{TR}_\text{Dec}^{resp}$), with  $M_{fused}$ as key and value, is employed:
\begin{flalign}
&O = \text{TR}_\text{Dec}^{resp}(E_W(R_{0:t-1}), M_{fused}, M_{fused}),\\
&\mathcal{P}_{resp} = \softmax(\FC_{decode}(O)),\\
&p(R_i|C, R_{0:i-1}) = \mathcal{P}_{resp}[i],
\end{flalign}
where $O\in \mathbb{R}^{t\times D_h}$, $t$ is the number tokens in response $R$ ($R_0$ is \texttt{<start>} token), $\FC_{decode}$ is a fully-connected layer of output size $|V|$ --- also the vocabulary size ---, $\mathcal{P}_{resp}\in \mathbb{R}^{t\times |V|}$, and  $p(R_i|C, R_{0:i-1})$ is the probability distribution on each response token.

Finally, categorical cross-entropy quantifies the generation loss with respect to the gold response $R_{gold}$:
\begin{flalign}
L_{resp} = - \log p(R_{gold}|C).
\end{flalign}

\subsection{Training}

Naturally, we combine all the losses for model training:
\begin{flalign}
\mathcal{L} = \alpha L_{cls} + \beta (L_{pos}^{ELBO} + L_{neg}^{ELBO}) + \gamma L_{resp}.
\end{flalign}
Total loss $\mathcal{L}$ is optimized using Adam~\cite{DBLP:journals/corr/KingmaB14} optimizer with learning-rate, patience, and batch-size set to 0.0001, 2, and 16, respectively.
Loss weights, $\alpha, \beta,$ and $\gamma$ are set to 1. For the sake of comparability with the SOTA, the semantic word embeddings ($E_W$) are initialized with GloVe~\cite{DBLP:conf/emnlp/PenningtonSM14} embeddings. All the hyper-parameters are optimized using grid search on the validation set, resulting $D_h$ and beam-size being 300 and 5, respectively.

\section{Experimental Settings}

During inference, we use the emotion classifier~(\cref{eqn:emo-classifier}) with emotion grouping~(\cref{tab:emo_group}) to determine the positivity or negativity of the context that is necessary for the mimicking and non-mimicking emotion representations.

\subsection{Dataset}

We evaluate our method on \textsc{EmpatheticDialogues}\footnote{\url{https://github.com/facebookresearch/EmpatheticDialogues}}~\cite{Rashkin2018IKT}, a dataset that contains 24,850 open-domain dyadic conversations between two users, where one responds emphatically to the other. For our experiments, we use the 8:1:1 train/validation/test split, defined by the authors of this dataset. Each sample consists of a context --- defined by an excerpt of a full conversation and the emotion of the user --- and the  empathetic response to the last utterance in the context. There are a total of 32 different emotion categories roughly uniformly distributed across the dataset.

\subsection{Baselines and State of the Art}

We do not compare MIME with affective response generation models~\cite{zhou2018emotional} as they require the response emotion to be explicitly provided, and the response may not necessarily be empathetic.
As such, MIME is compared against the following models:


\paragraph{Multitask-Transformer Network (Multi-TR).} Following \citet{Rashkin2018IKT}, a transformer encoder-decoder~\cite{NIPS2017_7181} generates a response as the user emotion is classified from the encoder output --- equivalent to $c$ in \cref{eqn:c}.

\paragraph{Mixture of Empathetic Listeners (MoEL).} This state-of-the-art method~\cite{lin-etal-2019-moel} performs user-emotion classification as Multi-TR. However, in contrast to our method, it employs emotion-specific decoders whose outputs are aggregated and fed to a final decoder to generate the empathetic response.

\subsection{Evaluation}

Although BLEU~\cite{papineni-etal-2002-bleu} has long been used to compare system-generated response against the human-gold response, \citet{liu-etal-2016-evaluate} argues against its efficacy in open-domain where the gold response is not necessarily the only correct response. Therefore, as MoEL, we keep BLEU mostly as reference. Following MoEL and \citet{Rashkin2018IKT}, we rely on human-evaluated metrics:

\paragraph{Human Ratings.} Firstly, we randomly sample four instances of each of the 32 emotion labels from the test set, resulting in a total of 128 instances, compared to the 100 instances used for the evaluation of MoEL. Next, we ask three human annotators to rate each sub-sampled model response on a scale from 1 to 5 (best score) on three distinct attributes: \textbf{empathy} (How much \emph{emotional understanding} does the response show?), \textbf{relevance} (How much \emph{topical relevance} does the response have to the context?), and \textbf{fluency} (How much \emph{linguistic clarity} does the response have?).
Scores across 128 samples and three annotators are averaged to obtain the final rating.

\paragraph{Human A/B Test.} Given two models A and B --- in our case MoEL and MIME (our model), respectively --- we ask three human annotators to pick the model with the best response for each of the 128 sub-sampled test instances. The annotators can select a \emph{Tie} if the responses from both models are deemed equal. The final verdict on each instance is determined by majority voting. In case no two annotators agree on a selection -- that is all three annotators reached three distinct conclusions: MoEL, MIME, and Tie -- we bring in a fourth annotator. From this, we calculate the percentage of samples where A or B generates the better response and where A and B are equal.

\section{Results and Discussions}
\label{sec:results}

\subsection{Response-Generation Performance}
\label{sec:response-generation-performance}

\begin{table}[ht]
    \centering
    \small
    \resizebox{\linewidth}{!}{
    \begin{tabular}{l||c|c|ccc}
    \hline
         \multirow{2}{*}{Methods} & \multirow{2}{*}{\#params.} & \multirow{2}{*}{BLEU} & \multicolumn{3}{c}{Human Ratings}\\
         \cline{4-6} & & & Emp. & Rel. & Flu. \\
         \hline
         \hline
         Multi-TR & 16.95M & 2.92 & 3.67 & 3.47 & 4.30 \\
         MoEL (SOTA) & 23.10M & 2.90 & 3.71 & 3.32 & \bf 4.31 \\
         MIME & 17.80M & 2.98 & \bf 3.87 & \bf 3.60 & 4.28 \\
         \hline
    \end{tabular}
    }
    \caption{Comparison among MIME (our model), baselines, and the state-of-the-art MoEL; Emp., Rel., and Flu. stand for Empathy, Relevance, and Fluency, respectively; the best score for each metric is highlighted by bold font.}
    \label{tab:overall-results}
\end{table}

Following \cref{tab:overall-results}, responses from MIME show improved \emph{empathy} over MoEL and Multi-TR. We surmise this was achieved by modeling our primary intuition of appropriately mimicking user's emotion in the context thorough stochasticity and positive/negative grouping. Moreover, the usage of trained emotion embeddings ($E_\mathcal{E}$), shared between the emotion classifier and response decoder, seems to encode refined context-invariant emotional and emotion-specific linguistics cues that may lead to empathetically-improved response generation. The SOTA model, MoEL, does train a similar emotion embedding, but it is setup as the key of a key-value memory~\cite{miller-etal-2016-key} which leads to a weaker connection with the decoder, resulting in less emotional-context flow. We believe this embedding sharing further leads to improved \emph{relevance} rating for MIME, since contextual information flow is now shared between emotion embeddings and encoder output (\cref{eqn:H}). This sharing intuitively leads to refinement in context flow.

However, we also observe that the responses from our model have worse \emph{fluency} than the other models, including MoEL. This might be attributed to the very structure of the decoder, that seems to refine emotional context well. This may have coerced the final transformer decoder to focus more on emotionally-apt tokens of the response than appropriate stop-words that have no intrinsic emotional content, but lead to grammatical clarity.

\paragraph{Human A/B Test.} Based on the results in \cref{tab:a-b}, we note that the responses from MIME are more often preferable to humans than the responses from MoEL and Multi-TR. This correlates with the results in \cref{tab:overall-results} that indicate better empathy and contextual relevance for MIME.
Further, the annotators prefer the responses from MIME with stochasticity (STC) than otherwise. \cref{tab:with-STC vs without-STC} shows the impact of stochasticity on the responses.

\begin{table}[t]
    \centering
    \small
    \begin{tabular}{l||ccc}
         \hline
         \stack{MIME}{vs.} & \stack{MIME}{Wins} & \stack{MIME}{Loses} & Tie \\
         \hline
         \hline
         Multi-TR & 42.25\% & 24.60\% & 33.15\%\\
         MoEL & 38.82\% & 28.32\% & 32.86\% \\
         MIME w/o STC & 39.84\% & 23.43\% & 36.73\%\\
         \hline
    \end{tabular}
    \caption{Human A/B test results for MIME vs. MIME without stochasticity (STC), MoEL, and Multi-TR.}
    \label{tab:a-b}
\end{table}

\paragraph{Performance on Positive and Negative User Emotions.}

\begin{table}[ht!]
    \centering
    \small
    \resizebox{\linewidth}{!}{
    \begin{tabular}{l||cc|cc|cc}
        \hline
        \multirow{2}{*}{Ratings} & \multicolumn{2}{c|}{Multi-TR} & \multicolumn{2}{c|}{MoEL} & \multicolumn{2}{c}{MIME} \\
         & Pos & Neg & Pos & Neg & Pos & Neg\\
        \hline
        \hline
         Empathy & 3.77 & 3.61 & 3.73 & 3.76 & \bf 4.00 & \bf 3.80\\
         Relevance & 3.51 & 3.45 & 3.21 & 3.40 & \bf 3.77 & \bf 3.49\\
         Fluency & 4.33 & 4.28 & \bf 4.35 & \bf 4.30 & 4.33 & 4.26\\
        \hline
    \end{tabular}
    }
    \caption{Models performance on positive and negative context; Pos and Neg stand for positive and negative emotions, respectively; the best score for each metric-polarity combination is highlighted by bold font.}
    \label{tab:pos-neg}
\end{table}
\begin{figure*}[t]
    \centering
    \begin{subfigure}{0.45\linewidth}
        \centering
        \includegraphics[width=0.9\linewidth]{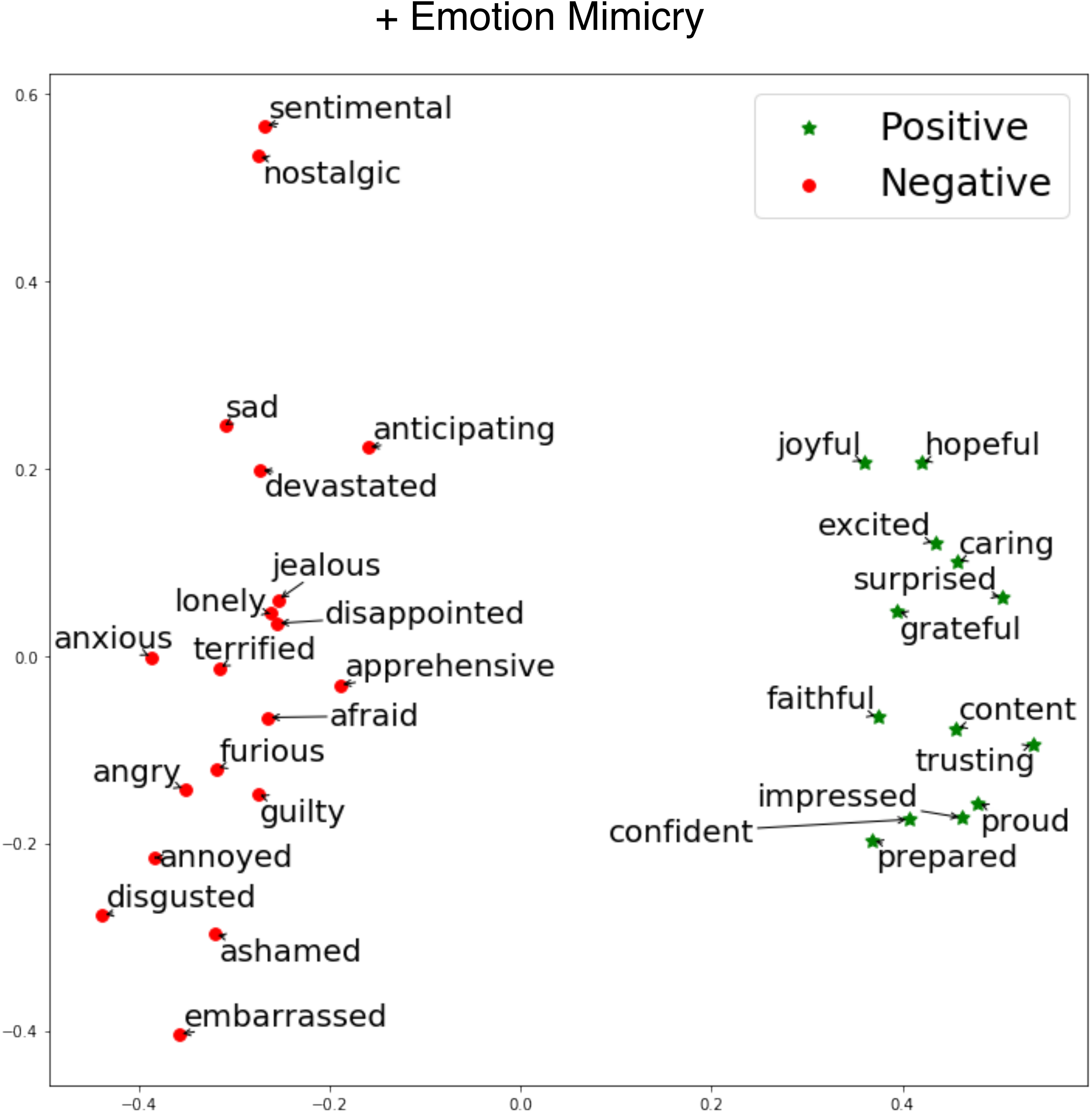}
        \caption{}
        \label{fig:mimic-plot}
    \end{subfigure}
    ~
    \begin{subfigure}{0.45\linewidth}
        \centering
        \includegraphics[width=0.9\linewidth]{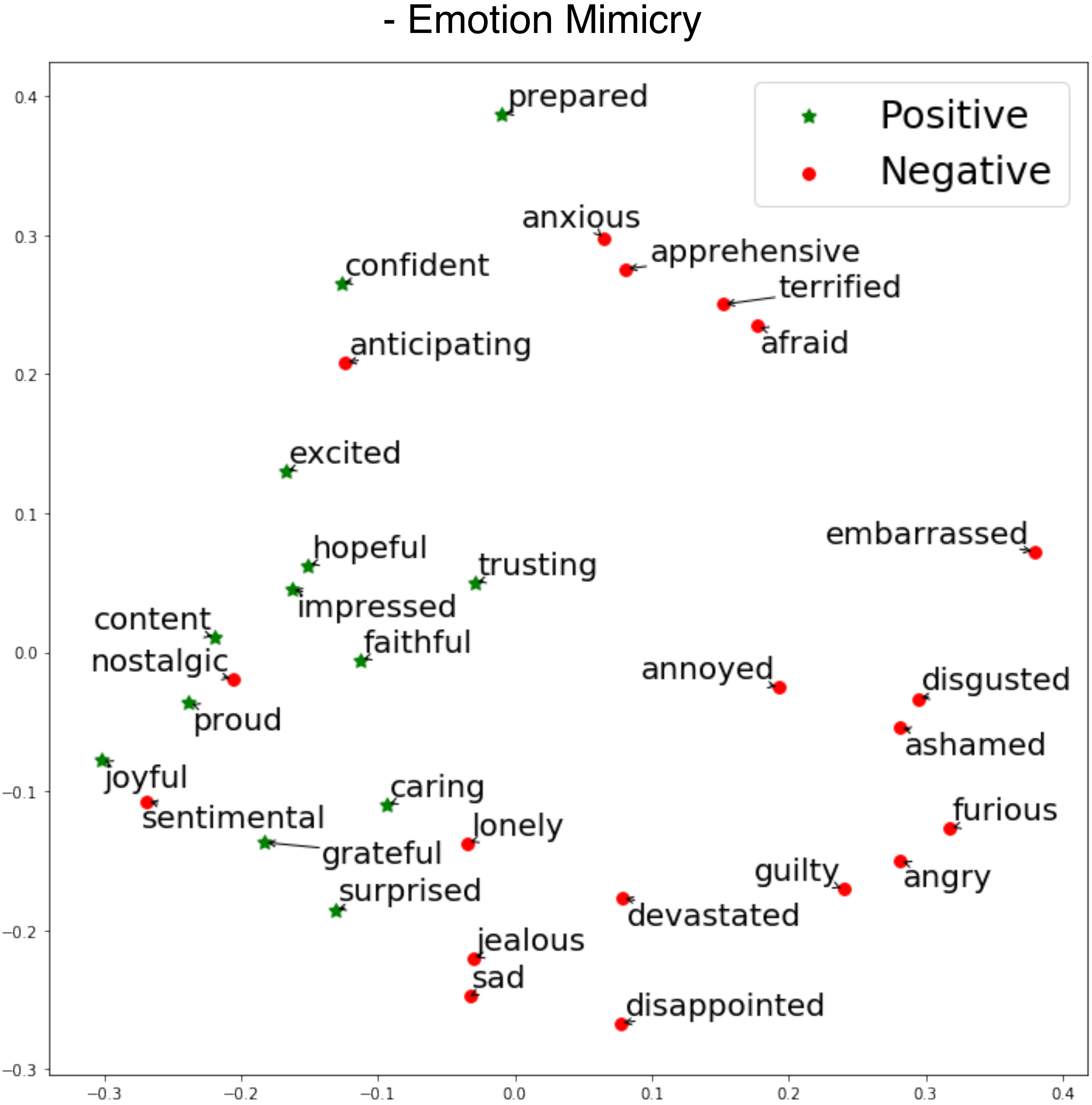}
        \caption{}
        \label{fig:no-mimic-plot}
    \end{subfigure}
    \caption{(a) and (b) plot the emotion embeddings, with and without emotion mimicry, respectively, mapped to two dimensions using top-two principal components.}
\vskip -0.1in
    \label{fig:MvsnoM}
\end{figure*}
We observe (\cref{tab:pos-neg}) that the responses generated by MIME for both positive and negative user emotions are generally better in terms of empathy and fluency. Interestingly, MoEL seems to perform better on responding to negative emotions than to positive emotions in terms of empathy and fluency. We posit this stems from the abundance of negative samples in the dataset as compared to positive samples --- 13 positive and 19 negative emotions roughly uniformly distributed. This may suggest that MoEL is more sensitive to positive/negative context imbalance in the dataset than MIME and Multi-TR.

\subsection{Ablation Study}

\begin{table}[ht!]
    \centering
    \small
    \resizebox{\linewidth}{!}{
    \begin{tabular}{cc||c|ccc}
        \hline
        Emotion & Emotion & \multirow{2}{*}{BLEU} & \multicolumn{3}{c}{Human Ratings} \\
        \cline{4-6} Mimicry & Grouping & & Emp. & Rel. & Flu.\\
        \hline
        \hline
        \xmark & \xmark & 2.45 $\pm$ 0.01 & 3.14 & 3.58 & 4.23 \\
        \xmark & \cmark & 2.96 $\pm$ 0.02 & 3.67 & \bf 3.63 & 4.09 \\
        \hline
        \cmark & \cmark & 2.98 $\pm$ 0.01 & \bf 3.87 & 3.60 & \bf 4.28 \\
        \hline
    \end{tabular}
    }
    \caption{Results of ablation; Emp., Rel., and Flu. stand for Empathy, Relevance, and Fluency, respectively.}
    \label{tab:ablation}
\end{table}

\paragraph{Effect of Emotion Mimicry.}
To assess the contribution of user-emotion mimicry, we disabled it by passing $e_g$ (\cref{eqn:e_g}) directly to \cref{eqn:Hm,eqn:Hnm}. This results in a substantial drop in \emph{empathy}, by 0.2 as per \cref{tab:ablation}. We delve deeper by plotting the emotion embeddings produced with and without emotion mimicry in \cref{fig:mimic-plot} and \cref{fig:no-mimic-plot}, respectively. It is evident that the separation of positive and negative emotions clusters is much clearer with emotion mimicry than otherwise, suggesting better emotion understanding in the prior case through emotion disentanglement. On the other hand, we observe slight increase of \emph{relevance}, by 0.03. We surmise this is caused by the absence of the confounding effect of swapping the value of $m$ and $\widetilde{m}$, in \cref{eqn:m,eqn:nm}, depending on the user emotion type. This may coerce the same set of parameters to learn processing both positive and negative emotions.
\begin{table*}[t]
\centering
\small
\resizebox{\textwidth}{!}{%
\begin{tabular}{|p{0.08\textwidth}|p{0.92\textwidth}|}
\hline

\textbf{Label} & 
lonely \\ 
\textbf{Context} & 
\textbf{Speaker}: i there , dont know what to do , jst broke up with my girlfirned , we were 8 years together \\ 
\textbf{MIME} &
oh no ! what happened ? \\ 
\textbf{MoEL} & 
oh wow , that must have been very exciting ! \\ 
\textbf{Gold} &
sorry to hear ! do you have any idea about the break up ? did you think about it ?\\
\hline

\textbf{Label} &
excited \\ 
\textbf{Context} & 
\textbf{Speaker}: i have a huge chance to win a ps4 ! \\ 
\textbf{MIME} &
that is great ! are you excited ? \\ 
\textbf{MoEL} &
that sounds like a lot of fun   \\  
\textbf{Gold} &
wow , please make it real . \\ 
\hline

\textbf{Label} &
faithful \\ 
\textbf{Context} & 
\textbf{Speaker}: there was this one time i was fairly drunk at a bar and this attractive girl was trying to make moves on me , but i did my best to ignore it all.. having a girlfriend and all . \\
\textbf{MIME} &
wow , that must have been so \textbf{embarrassing} . \\ 
\textbf{MoEL} &
that is terrible, did you confront her ? \\ 
\textbf{Gold} &
that is good ! you should be proud of yourself \\ 
\hline

\textbf{Label} &
disappointed \\ 
\textbf{Context} & 
\textbf{Speaker}: i lost my favorite hoodie somewhere . i am so upset . like how could i lose a hoodie ? it is so big , did i just leave it somewhere ? \\ 
\textbf{MIME} &
well that is not good , i hope you have to be careful next time \\ 
\textbf{MoEL} &
i am sorry to hear that . i am sorry to hear that . i am sorry to hear that .\\ 
\textbf{Gold} &
did you retrace your steps \\ 
\hline



\end{tabular}%
}
\caption{Comparison between MIME (\emph{with STC}) and MoEL.}
\label{tab:MIME vs MoEL}
\end{table*}
\paragraph{Effect of Emotion Grouping.}
Looking at \cref{tab:ablation}, we observe a performance drop in both \emph{empathy} and \emph{relevance}, by 0.73 and 0.02, respectively, in the absence of emotion grouping. This indicates the importance of having positive and negative emotions treated separately, rather than huddling them into a single distribution. We posit that the latter case causes all the emotions to compete for importance which may lead to emotion uniformity in some cases or one emotion-type overwhelming the other in other cases. This in turn may lead to emotionally mundane and generic responses.



\subsection{Case Study}
\begin{figure}[t]
    \centering
    \includegraphics[width=0.9\linewidth]{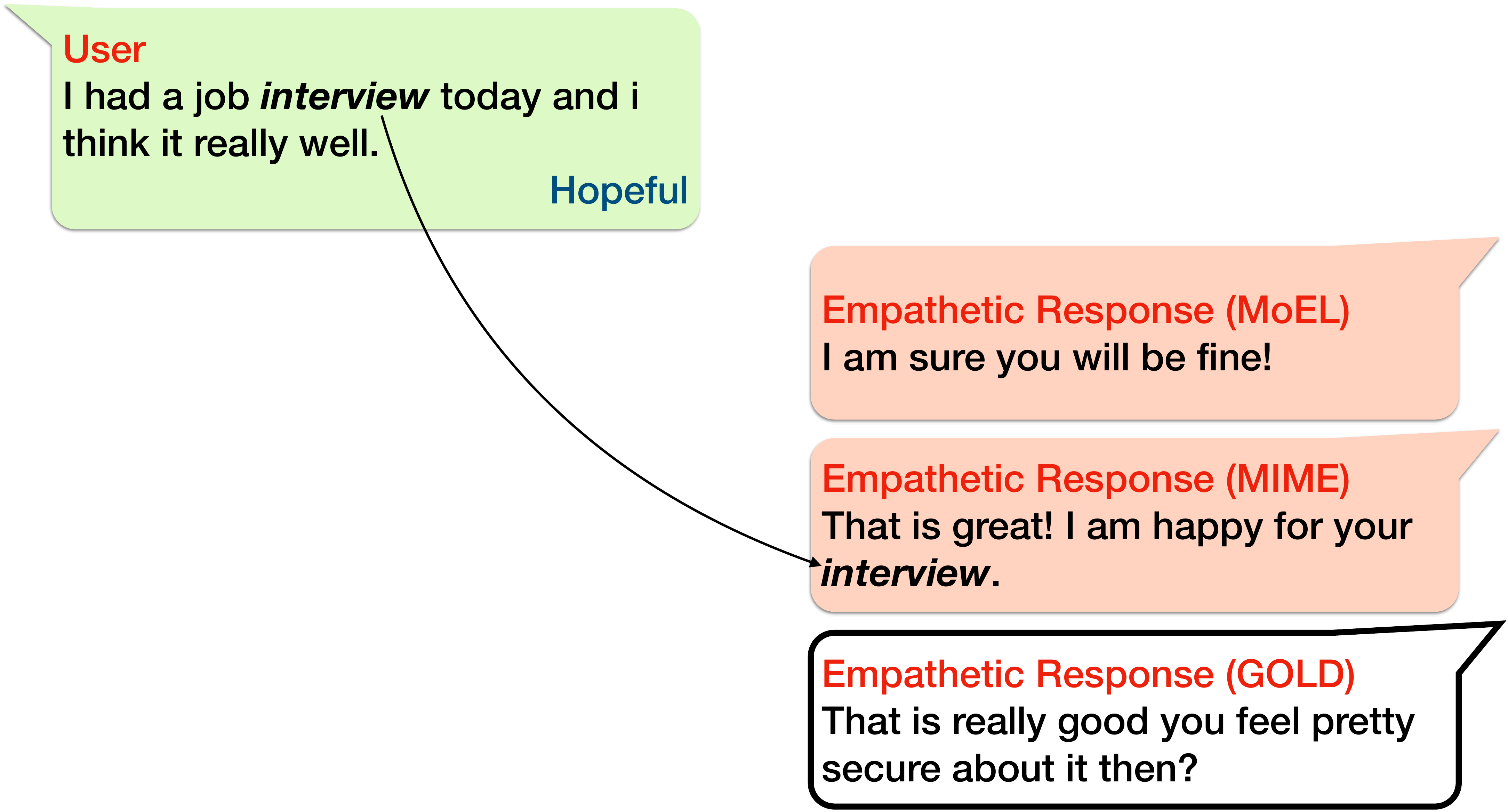}
    \caption{A test sample where MIME responds with key information from the context.}
    \label{fig:case-mime-better1}
\end{figure}

\paragraph{Context Capturing.}
Based on the comparative results for \emph{relevance} shown in \cref{tab:overall-results}, MIME appears to generate responses that are a closer fit to the context than MoEL does. \cref{fig:case-mime-better1} shows a test instance where MIME pulls key information from the context --- the word `interview' --- to generate an empathetic and relevant response. The response from MoEL is also empathetic, but somehow more generic. We surmise that this can be attributed to the two-way context flow through the emotion embedding sharing and encoder output, as discussed in \cref{sec:response-generation-performance}.
\begin{figure}[ht]
    \centering
    \includegraphics[width=0.9\linewidth]{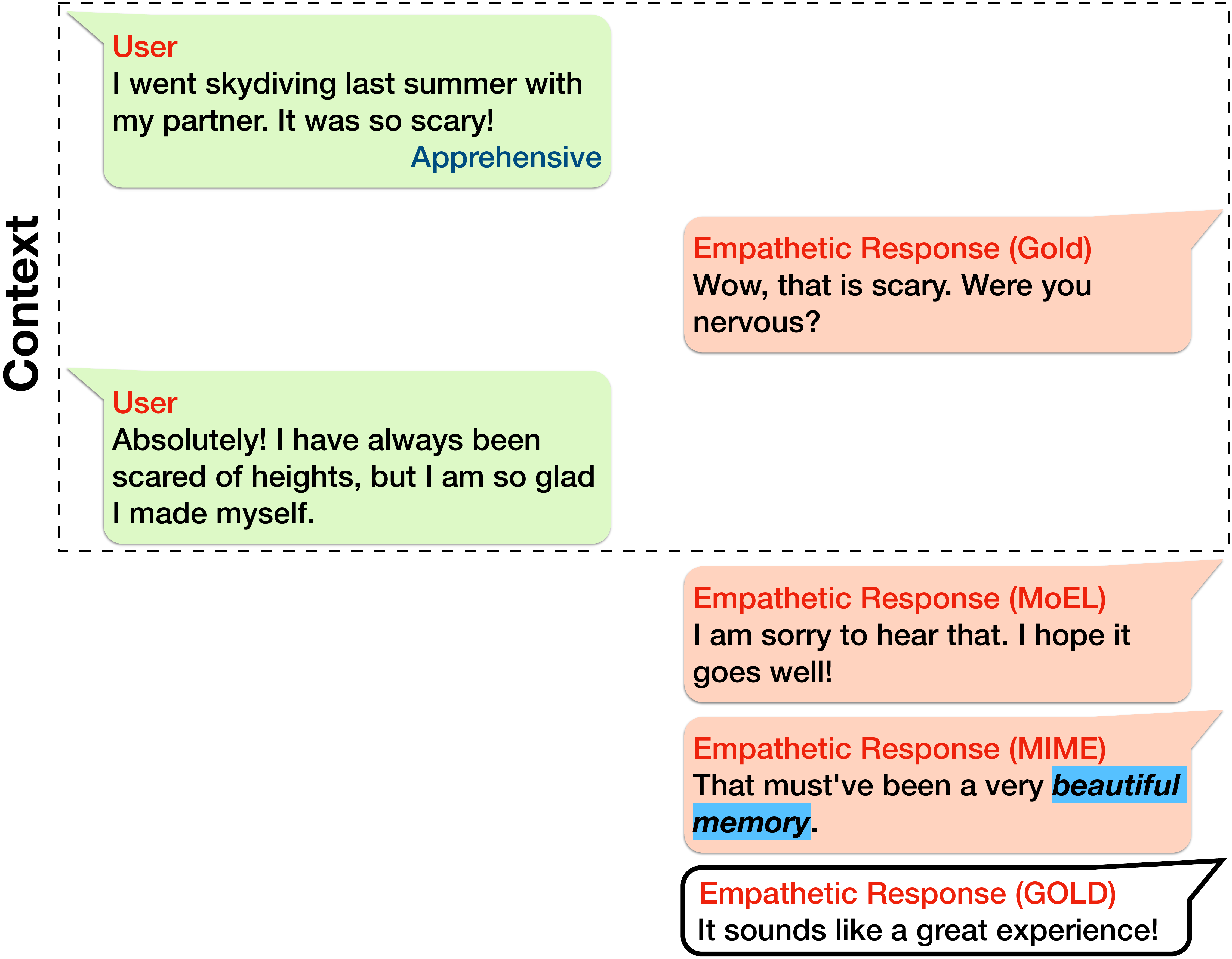}
    \caption{A test sample where MIME responds with subtle information from the context.}
\label{fig:case-mime-better2}
\end{figure}

Similarly, \cref{fig:case-mime-better2} shows a conversation with an \emph{apprehensive} user who shares a frightening story with a positive outcome. Here, MoEL fixates on the initial negative emotion of the user and replies with an unwarranted negatively empathetic response. MIME, however, responds with appropriate positivity hinted at the last utterance. Moreover, it is able to correctly interpret the events described as a `beautiful memory', which is truly empathetic and relevant. Again, strong mixture of context and emotion, facilitated by the emotion embedding sharing, is likely to be responsible for this. We show more examples generated by both MoEL and MIME in \cref{tab:MIME vs MoEL}.


\subsection{Error Analysis}

\begin{figure}[ht!]
    \centering
    \includegraphics[width=\linewidth]{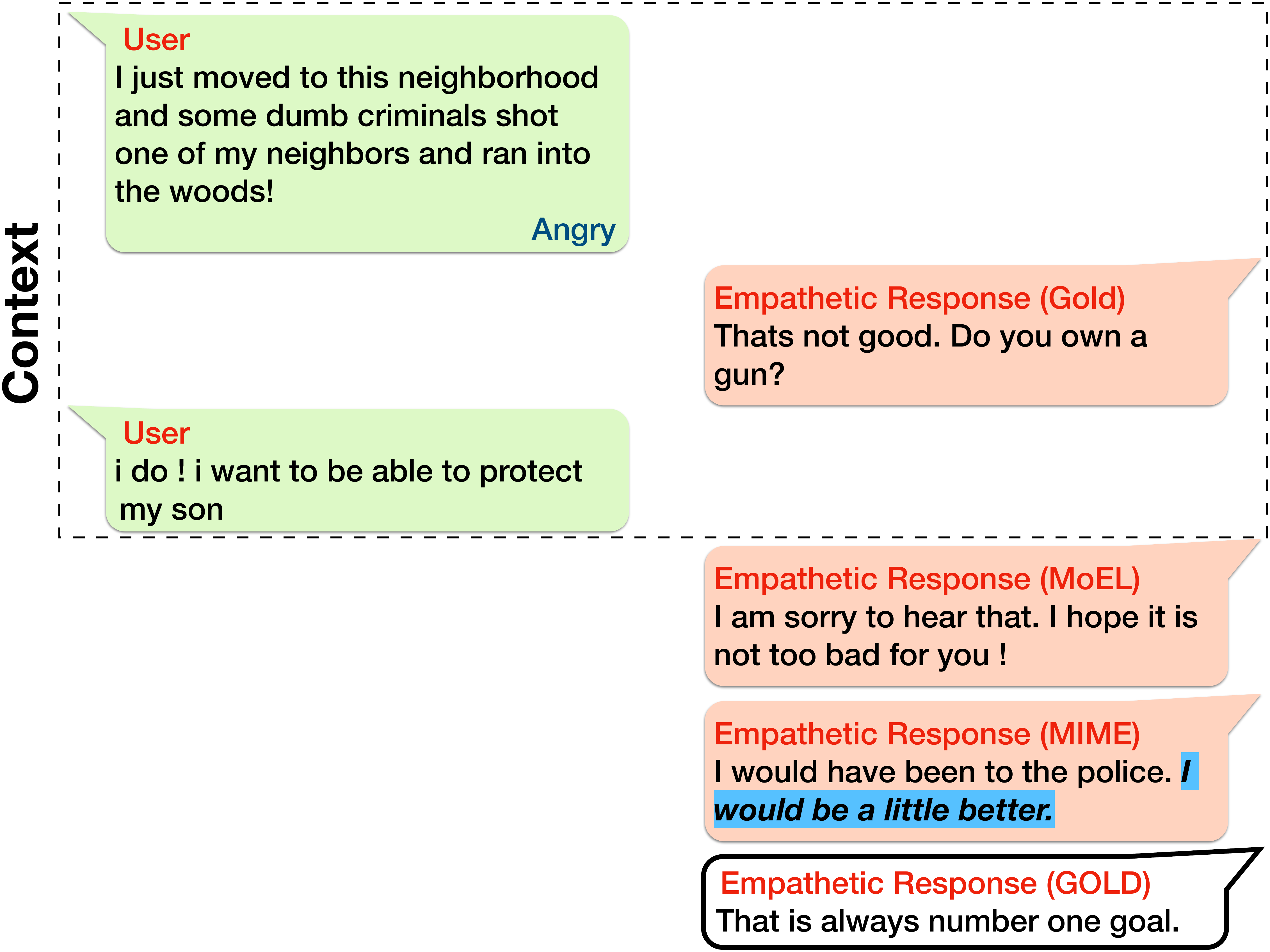}
 \caption{A test sample where MIME responds with a malformed utterance.}
\vskip -0.1in
\label{fig:case-mime-worse1}
\end{figure}

\begin{table*}[ht!]
\centering
\small
\resizebox{\textwidth}{!}{%
\begin{tabular}{|p{0.08\textwidth}|p{0.92\textwidth}|}
\hline


  
\textbf{Label} &
  anticipating \\ 
\textbf{Context} &
\textbf{Speaker}: i am looking forward to going on vacation in a few weeks ! we   have a condo reserved on the beach , with fantastic ocean views . i am ready   ! \\ 
\textbf{with \emph{STC}} &
  that is \textbf{awesome} ! i have been there . i hope you have a great time ! \\ 
\textbf{no \emph{STC}} &
  that is great ! i have never been there . \\ 
\textbf{Gold} &
  ah , that sounds fantastic ! which ocean will you be enjoying ? \\ \hline
  
\textbf{Label} &
  jealous \\ 
\textbf{Context} &
  \textbf{Speaker}:  my friend is a surgeon and we were discussing salary .  he easily   makes 200,000 a year and he is only 32. it is crazy and i was jealous of him   . \\ 
\textbf{with \emph{STC}} &
  i would be so \textbf{jealous} ! \\ 
\textbf{no \emph{STC}} &
  that is a good idea . \\ 
\textbf{Gold} &
  you should n't , because that is a very stressful job \\ \hline
  
\textbf{Label} &
  proud \\ 
\textbf{Context} &
  \textbf{Speaker}: my son graduated . \\ 
\textbf{with \emph{STC}} &
  \textbf{congrats} ! that is a \textbf{great accomplishment} ! \\ 
\textbf{no \emph{STC}} &
  that is great ! how old is he ? \\ 
\textbf{Gold} &
  from where ? \\ \hline

\end{tabular}%
}
\caption{Comparison between some responses from MIME with and without stochasticity (\emph{STC}).}
\label{tab:with-STC vs without-STC}
\end{table*}

\paragraph{Low Fluency.} As evidenced by \cref{tab:overall-results}, MIME falters in \emph{fluency} as compared to MoEL. \cref{fig:case-mime-worse1} shows an instance where MoEL generates an empathetic, yet somewhat generic, and fluent response. In contrast, the first response utterance from MIME --- ``\emph{I would have been to the police}'' --- does make  contextual sense. However, the second utterance ``\emph{I would be a little better}'' reads incoherent and semantically unclear. Perhaps the model meant something like `I would have felt a little safer'. We repeatedly observed such errors, leading to poor fluency. Given the empathy- and relevance-focused structure of our model, we think MIME focused on learning empathy and relevance, at the cost of fluency. We believe this issue could be mitigated with additional training samples.

\paragraph{Response to Surprised User Context.}
In our experiments, we assumed the emotion \emph{surprised} to be positive~(\cref{tab:emo_group}), and thus MIME responds with positivity to most test instances incurring \emph{surprise} as a user emotion. However, this is not accurate, as one can be both positively and negative surprised --- ``\emph{I recently found out that a person I [...] admired did not feel the same way. I was pretty surprised''} vs ``\emph{This mother's day was amazing!}''.
We posit that re-annotating the instances with a negatively-surprised user with a new negative emotion, namely \emph{shocked}, should help alleviate this issue significantly.

\paragraph{Emotion Classification.}
The \{top-1, top-2, top-5\} emotion-classification accuracies for MoEL are \{38\%, 63\%, 74\%\}, as compared to \{34\%, 58\%, 77\%\} for MIME. Since the emotion embeddings are shared between encoder and decoder in MIME, it supposedly also encodes some generation-specific information in addition to pure emotion as discussed in \cref{sec:response-generation-performance}, thereby hindering the overall emotion-classification performance. Notably, MIME also performs well on top-5 classification. This is likely due to MIME's ability to discern positive and negative emotion types --- as indicated by \cref{fig:mimic-plot} --- that comes into prominence as you add more likely-labels into the consideration of top-$k$ classification by raising $k$.


\section{Conclusion}

This paper introduced a novel empathetic generation strategy that relies on two key ideas: emotion grouping and emotion mimicry. Also, stochasticity was applied to the emotion mixture for varied response generation. We have shown through several human evaluations and ablation studies that our model is better equipped for empathetic response generation than  existing models. However, there remains much room for improvement, particularly in terms of \emph{fluency} where our model falters. Moreover, emotions like `surprise' and `anticipation' might be explicitly dealt with due to their ambiguous polarity.

 \section*{Acknowledgements}
 This research is supported by A*STAR under its RIE 2020 Advanced Manufacturing and Engineering (AME) programmatic grant RGAST2003 
, and by the John Templeton Foundation (grant \#61156). Any opinions, findings, and conclusions or recommendations expressed in this material are those of the authors and do not necessarily reflect the views of A*STAR, the National Science Foundation, or the John Templeton Foundation. 


\bibliographystyle{acl_natbib}
\bibliography{coling2020,rada}

\end{document}



\begin{table*}[!b]
\centering
\resizebox{\textwidth}{!}{%
\begin{tabular}{|p{0.1\textwidth}|p{0.9\textwidth}|}
\hline

\textbf{Label} &
  sad \\ \hline
\textbf{Context} &
  \begin{tabular}{p{0.85\textwidth}}
  \textbf{Speaker}:    one of the saddest things to me is when people underestimate   what they can do  and/or are capable of  \\ \textbf{Listener}:   i know ,   that actually happens to me sometimes !   last year i did not get accepted into   this fellowship and for a couple of weeks i was underestimating myself a   lot \\ \textbf{Speaker}:  that is perfectly natural .  you sound like the kind of person , though , that quickly  regained their bearings . disappointments like that usually come with some   *good* lessons .\end{tabular} \\ \hline
\textbf{with \emph{STC}} &
  that is a \textbf{great attitude} to have . i am sure you will do great ! \\ \hline
\textbf{no \emph{STC}} &
  i am glad that you are able to   do that ! \\ \hline
\textbf{Gold} &
  yea for sure , i definitely learned that sometimes life just works in a funny way .  when 's the last time you have been really disappointed ? \\ \hline
  \hline
  \hline
  
\textbf{Label} &
  anticipating \\ \hline
\textbf{Context} &
  \textbf{Speaker}: i am looking forward to going on vacation in a few weeks ! we   have a condo reserved on the beach , with fantastic ocean views . i am ready   ! \\ \hline
\textbf{with \emph{STC}} &
  that is \textbf{awesome} ! i have been there . i hope you have a great time ! \\ \hline
\textbf{no \emph{STC}} &
  that is great ! i have never been there . \\ \hline
\textbf{Gold} &
  ah , that sounds fantastic ! which ocean will you be enjoying ? \\ \hline
  \hline
  \hline
  
\textbf{Label} &
  jealous \\ \hline
\textbf{Context} &
  \textbf{Speaker}:  my friend is a surgeon and we were discussing salary .  he easily   makes 200,000 a year and he is only 32. it is crazy and i was jealous of him   . \\ \hline
\textbf{with \emph{STC}} &
  i would be so \textbf{jealous} ! \\ \hline
\textbf{no \emph{STC}} &
  that is a good idea . \\ \hline
\textbf{Gold} &
  you should n't , because that is a very stressful job \\ \hline
  \hline
  \hline
  
\textbf{Label} &
  proud \\ \hline
\textbf{Context} &
  \textbf{Speaker}: my son graduated . \\ \hline
\textbf{with \emph{STC}} &
  \textbf{congrats} ! that is a \textbf{great accomplishment} ! \\ \hline
\textbf{no \emph{STC}} &
  that is great ! how old is he ? \\ \hline
\textbf{Gold} &
  from where ? \\ \hline
  \hline
  \hline
  
\textbf{Label} &
  sad \\ \hline
\textbf{Context} &
\begin{tabular}{p{0.85\textwidth}} \textbf{Speaker}: i was just thinking   about how a few years ago , we lost our cat .\\ \textbf{Listener}: oh no ! what happened to it ?\\ \textbf{Speaker}: it was the first time he had gotten outside and he did not come   back that night . he was hit by a car . : (\end{tabular} \\ \hline
\textbf{with \emph{STC}} &
  oh wow ! that is really \textbf{scary} ! \\ \hline
\textbf{no \emph{STC}} &
  that is great ! i am glad you had to go through that . \\ \hline
\textbf{Gold} &
  so sad ! i am sorry . do not feel too bad its natural for them to want to   be outside . \\ \hline
  \hline
  \hline
  
\textbf{Label} &
  Lonely \\ \hline
\textbf{Context} &
  \begin{tabular}{p{0.85\textwidth}} \textbf{Speaker}: having the whole house   to your self sometimes is not as good as it may seem !\\ \textbf{Listener}: i can imagine the loneliness someone can experience\\ \textbf{Speaker}: my whole family is out the house for a few months . i am going to   be so bored with no one around ...\end{tabular} \\ \hline
\textbf{with \emph{STC}} &
  it is really \textbf{hard} . i hope you can find some \textbf{new friends} . \\ \hline
\textbf{no \emph{STC}} &
  it is a great feeling to have a new one ! \\ \hline
\textbf{Gold} &
  get your self a friend to stay with for a couple of months . \\ \hline
  \hline
  \hline

\textbf{Label} &
  guilty \\ \hline
\textbf{Context} &
  \textbf{Speaker}: yeah about 10 years ago i had a horrifying experience . it was   100 \% their fault but they hit the water barrels and survived . they had no   injuries but they almost ran me off the road \\ \hline
\textbf{with \emph{STC}} &
  oh no ! that is \textbf{scary} ! did you have \textbf{insurance}? \\ \hline
\textbf{no \emph{STC}} &
  oh no ! i am so sorry to hear that . \\ \hline
\textbf{Gold} &
  did you suffer any injuries ? \\ \hline

\end{tabular}%
}
\caption{Comparison between some responses from MIME with and without stochasticity (\emph{STC}).}
\label{tab:with-STC vs without-STC}
\end{table*}


\begin{table*}[!b]
\centering
\resizebox{\textwidth}{!}{%
\begin{tabular}{|p{0.1\textwidth}|p{0.9\textwidth}|}
\hline

\textbf{Label} & 
lonely \\ \hline
\textbf{Context} & 
\textbf{Speaker}: i there , dont know what to do , jst broke up with my girlfirned , we were 8 years together \\ \hline
\textbf{MIME} &
oh no ! what happened ? \\ \hline
\textbf{MoEL} & 
oh wow , that must have been very exciting ! \\ \hline
\textbf{Gold} &
sorry to hear ! do you have any idea about the break up ? did you think about it ?\\ \hline
\hline
\hline

\textbf{Label} &
excited \\ \hline
\textbf{Context} & 
\textbf{Speaker}: i have a huge chance to win a ps4 ! \\ \hline
\textbf{MIME} &
that is great ! are you excited ? \\ \hline
\textbf{MoEL} &
that sounds like a lot of fun   \\  \hline
\textbf{Gold} &
wow , please make it real . \\ \hline
\hline
\hline

\textbf{Label} &
faithful \\ \hline
\textbf{Context} & 
\textbf{Speaker}: there was this one time i was fairly drunk at a bar and this attractive girl was trying to make moves on me , but i did my best to ignore it all.. having a girlfriend and all . \\ \hline
\textbf{MIME} &
wow , that must have been so \textbf{embarrassing} . \\ \hline
\textbf{MoEL} &
that is terrible, did you confront her ? \\ \hline
\textbf{Gold} &
that is good ! you should be proud of yourself \\ \hline
\hline
\hline

\textbf{Label} &
disappointed \\ \hline
\textbf{Context} & 
\textbf{Speaker}: i lost my favorite hoodie somewhere . i am so upset . like how could i lose a hoodie ? it is so big , did i just leave it somewhere ? \\ \hline
\textbf{MIME} &
well that is not good , i hope you have to be careful next time \\ \hline
\textbf{MoEL} &
i am sorry to hear that . i am sorry to hear that . i am sorry to hear that .\\ \hline
\textbf{Gold} &
did you retrace your steps \\ \hline
\hline
\hline

\textbf{Label} &
  faithful \\ \hline
\textbf{Context} &
  \textbf{Speaker}: my friend 's boyfriend recently made a pass at me . i am married , and i also am really good friend 's with his girlfriend so i felt really bad . i made sure he knew i was loyal and honest to both ! \\ \hline
\textbf{MIME} &
 that is really nice of you . i am sure you will be fine . \\ \hline
\textbf{MoEL} &
 that is good . i have been married a lot of people like that . \\ \hline
\textbf{Gold} &
  thats horrible . did you tell on him ? \\ \hline
  \hline
  \hline

\textbf{Label} &
  sad \\ \hline
\textbf{Context} &
\begin{tabular}{p{0.85\textwidth}}
\textbf{Speaker}: i broke up with my girlfriend last week and it was really hard \\ \textbf{Listener}: oh no i am sorry to hear about that , were you together long ? \\ \textbf{Speaker}: yeah we where together for about 2 years but im getting over it now .\end{tabular} \\ \hline
\textbf{MIME} &
that is a good idea . i am sure you will be able to find a way out and make sure you find someone else . \\ \hline
\textbf{MoEL} &
 i am so sorry to hear that . i hope you are feeling better about it . \\ \hline
\textbf{Gold} &
that is good make yourself available and open for others and you will find the next relationship \\ \hline

\end{tabular}%
}
\caption{Comparison between MIME (\emph{with STC}) and MoEL.}
\label{tab:MIME vs MoEL}
\end{table*}